\documentclass{article}
\usepackage{amsmath,graphicx,mlspconf}
\usepackage{xcolor}
\usepackage{subcaption}

\usepackage{float} 

\usepackage{amsmath,amsfonts,bm}

\def\eqref#1{(\ref{#1})}

\def\1{\bm{1}}

\def\vf{{\bm{f}}}

\def\vw{{\bm{w}}}
\def\vx{{\bm{x}}}

\def\mA{{\bm{A}}}

\def\mF{{\bm{F}}}

\def\mH{{\bm{H}}}
\def\mI{{\bm{I}}}

\def\mK{{\bm{K}}}

\def\mQ{{\bm{Q}}}

\def\mV{{\bm{V}}}

\def\mZ{{\bm{Z}}}

\DeclareMathAlphabet{\mathsfit}{\encodingdefault}{\sfdefault}{m}{sl}
\SetMathAlphabet{\mathsfit}{bold}{\encodingdefault}{\sfdefault}{bx}{n}

\newcommand{\E}{\mathbb{E}}

\newcommand{\R}{\mathbb{R}}

\DeclareMathOperator*{\argmin}{arg\,min}

\def\bxi{{\boldsymbol{\xi}}}

\def\bGamma{{\boldsymbol{\Gamma}}}

\usepackage{amsthm}

\theoremstyle{plain}
\newtheorem{theorem}{Theorem}

\newtheorem{lemma}[theorem]{Lemma}
\newtheorem{corollary}[theorem]{Corollary}
\theoremstyle{definition}

\theoremstyle{remark}

\copyrightnotice{979-8-3503-2411-2/25/\$31.00 {\copyright}2025 IEEE}

\toappear{2025 IEEE International Workshop on Machine Learning for Signal Processing, Aug.\ 31-- Sep.\ 3, 2025, Istanbul, Turkey}

\title{Asymptotic Study of In-context Learning with Random Transformers through Equivalent Models}

\name{Samet Demir,\; Zafer Doğan\thanks{We acknowledge that this work is supported partially by TÜBİTAK under project 124E063 in ARDEB 1001 program. S.D. is supported by an AI Fellowship provided by KUIS AI Research Center and a PhD Scholarship (BİDEB 2211) from TÜBİTAK. The corresponding author is Zafer Doğan (zdogan@ku.edu.tr).}}
\address{
    MLIP Research Group, KUIS AI Center \& Department of EEE, Koç University\\
    İstanbul, Turkey
}

\begin{document}

\maketitle

\begin{abstract}
We study the in-context learning (ICL) capabilities of pretrained Transformers in the setting of nonlinear regression. Specifically, we focus on a random Transformer with a nonlinear MLP head where the first layer is randomly initialized and fixed while the second layer is trained. Furthermore, we consider an asymptotic regime where the context length, input dimension, hidden dimension, number of training tasks, and number of training samples jointly grow. In this setting, we show that the random Transformer behaves equivalent to a finite-degree Hermite polynomial model in terms of ICL error. This equivalence is validated through simulations across varying activation functions, context lengths, hidden layer widths (revealing a double-descent phenomenon), and regularization settings. Our results offer theoretical and empirical insights into when and how MLP layers enhance ICL, and how nonlinearity and over-parameterization influence model performance.
\end{abstract}

\begin{keywords}
In-context learning, transformer, deep learning theory, high-dimensional asymptotics.
\end{keywords}

\section{Introduction}
\label{sec:intro}
Transformers \cite{vaswani2017attention} have become the backbone of deep learning in recent years. One of their emerging capabilities is the so-called in-context learning (ICL), which enables task adaptation from prompts without modifying internal parameters \cite{brown2020language}. Given the popularity and effectiveness of Transformers for ICL, a theoretical understanding of their ICL capabilities has become a subject of significant interest.

Due to the complexity of analyzing ICL in full generality, existing studies often consider simplified settings. In particular, many works focus on ICL in linear regression or classification tasks using Transformer models stripped down to attention-only architectures, omitting multi-layer perceptrons (MLPs) entirely \cite{akyurek2023what, pmlr-v202-von-oswald23a, wu2024how, zhang2024trained}. While recent studies have begun exploring the role of nonlinear MLPs in Transformer-based ICL \cite{li2024nonlinear, kim2024transformers, oko2024pretrained}, several limitations remain. For example, \cite{li2024nonlinear} focuses specifically on classification tasks with ReLU-based MLPs, without considering broader activation functions or regression tasks. Meanwhile, \cite{kim2024transformers, oko2024pretrained} analyze Transformer variants in which the MLP layers precede the attention mechanism—contrary to the original Transformer architecture \cite{vaswani2017attention}, where input embeddings first pass through attention blocks before being processed by MLPs. As a result, a comprehensive understanding of how MLPs affect ICL performance in standard Transformer architectures remains incomplete.

To address this gap, we investigate the ICL performance of Transformers with nonlinear MLPs for nonlinear regression tasks through the lens of asymptotic analysis. Our approach bridges two research areas: ICL by Transformers and the asymptotic theory of MLPs. For the latter, we draw on Gaussian universality results \cite{hu2022universality, demir2024random, demir2025asymptotic}, which show that certain classes of MLPs are equivalent in behavior to Gaussian models with matched moments. This equivalence enables us to characterize the influence of nonlinear MLPs on the ICL performance of Transformers.

Concretely, we consider a random Transformer model with linear attention and a nonlinear MLP, where the first layer weights of the MLP are randomly initialized and fixed while the second layer is fully trained. In this setting, we show that the Transformer is asymptotically equivalent to a finite-degree polynomial model in terms of in-context learning error. We also identify the conditions under which the presence of a nonlinear MLP leads to improved ICL performance over a Transformer without MLPs. Specifically, our findings indicate that MLPs enhance ICL when the following conditions are met: (i) the activation function is appropriately chosen, (ii) the context length is sufficiently large, and (iii) the hidden dimension is suitably selected or the model is properly regularized. Furthermore, our analysis reveals that the ICL error of Transformers with nonlinear MLPs may exhibit non-monotonic behavior—commonly referred to as the "double-descent phenomenon"—as a function of model complexity. This extends prior observations of double descent in MLPs \cite{nakkiranoptimal2021, demir2023optimal} to the ICL setting.

Overall, our contributions are as follows:
\begin{enumerate}
    \item We find a polynomial model that performs equivalent to the random Transformer with MLPs for ICL.
    \item We characterize the conditions under which the Transformer with nonlinear MLPs outperforms the Transformer without MLPs.
    \item We demonstrate that non-monotonic behavior appearing in the ICL error of the Transformer with MLP can be mitigated with proper regularization. 
\end{enumerate}

\section{Related Work: ICL with Transformers}
The seminal work of \cite{brown2020language} first highlighted Transformers’ capacity for in-context learning, sparking a surge of empirical and theoretical investigations. On the empirical side, studies such as \cite{wei2022emergent}, \cite{olsson2022context}, and \cite{schaeffer2023are} demonstrated that ICL abilities tend to emerge more prominently as model size increases, emphasizing its significance in large-scale AI systems. To better understand the mechanisms behind ICL, researchers have employed controlled synthetic benchmarks, most notably linear regression tasks, using Transformers as in \cite{zhang2024trained}, \cite{garg2022can}, and \cite{raventos2024pretraining}. These benchmarks enable precise analysis by isolating specific architectural and data-driven effects.

Theoretically, numerous works have proposed that Transformers implicitly acquire algorithmic structures during pretraining, which are subsequently applied during ICL \cite{bai2023transformers, li2023transformers, akyurek2023what, ahn2023transformers, pmlr-v202-von-oswald23a, mahankali2024one, fu2024transformers, zhang2024trained, li2024finegrained, park2024competition}. However, the exact nature of these learned algorithms remains an open question. To gain tractability, many of these studies analyze simplified Transformer models—especially those with linearized self-attention. In particular, \cite{zhang2024trained} and \cite{lu2024incontext, lu2025asymptotic} provide detailed generalization analysis for linear attention-based Transformers in synthetic ICL tasks.

More recently, the inclusion of nonlinear MLP components in Transformer architectures has attracted attention. For example, \cite{li2024nonlinear} investigated ICL in a classification setting with ReLU-based MLP, while \cite{kim2024transformers} and \cite{oko2024pretrained} examined models where the MLP layer precedes the attention mechanism. However, these works either restrict the choice of activation function and task type or depart from the standard Transformer architecture \cite{vaswani2017attention}, where attention layers precede MLPs.

Consequently, a comprehensive theoretical treatment of ICL in Transformers with nonlinear MLP heads—adhering to the original architecture and applicable to general task settings such as regression—remains absent from the literature. This is the gap that our work seeks to address.

\section{Setting}

In this work, we investigate the ICL capabilities of pretrained Transformer architectures on nonlinear regression problems. 

\noindent Specifically, given a sequence of paired examples
\[
\bigl(\vx_1, y_1\bigr),\,\bigl(\vx_2, y_2\bigr),\dots,\bigl(\vx_{\ell}, y_{\ell}\bigr),\,(\vx_{\ell+1}, \,?),
\]
with each input \(\vx_i \in \mathbb{R}^d\) and corresponding scalar response \(y_i \in \mathbb{R}\) drawn independently from an unknown joint distribution, the model must infer the underlying mapping from the first \(\ell\) examples and predict the label \(y_{\ell+1}\) for a new input \(\vx_{\ell+1}\). Here, \(\ell\) denotes the context length.

We posit a nonlinear relationship between \(\vx\) and \(y\) governed by a context-specific parameter vector \(\boldsymbol{\xi}\in\mathbb{R}^d\), subject to additive Gaussian noise. Although \(\boldsymbol{\xi}\) remains constant within a given context, it is resampled across different contexts, imposing the requirement that the model estimate \(\boldsymbol{\xi}\) from the observed pairs before generalizing to the new input. Formally, we generate data according to  
\begin{align}
  \vx_i \sim \mathcal{N}\bigl(\mathbf{0}, \mI_d/d\bigr), \quad  y_i = \sigma_*\bigl(\boldsymbol{\xi}^T \vx_i\bigr) + \epsilon_i,
  \label{eq:data_model}
\end{align}
where $\sigma_*: \R \to \R$ is a nonlinear function and $\epsilon_i\sim\mathcal{N}(0,\rho)$ for $\rho > 0$, while the task vector itself is drawn as  
\begin{align}
    \boldsymbol{\xi} \sim \mathcal{N}\bigl(\mathbf{0}, \mI_d\bigr).
\end{align}
This probabilistic formulation captures both the nonlinear dependency and the variability inherent in regression tasks. We then consider $k$ random task vectors $\bxi$ to be used to generate the training samples, while the in-context learning error is measured with an expectation over all possible task vectors.

To feed these examples into a Transformer, we follow the literature \cite{zhang2024trained} and form an embedding matrix \(\mZ\) by stacking the feature vectors atop their scalar labels (with a zero placeholder for the final label), namely  
\begin{align}
      \mZ \;=\;
  \begin{bmatrix}
    \vx_1 & \vx_2 & \cdots & \vx_{\ell} & \vx_{\ell+1} \\[3pt]
    y_1           & y_2           & \cdots & y_{\ell}           & 0
  \end{bmatrix} \in \mathbb{R}^{(d+1)\times (\ell+1)}.
  \label{eq:embedding_matrix}
\end{align}

The Transformer must leverage this representation to infer the underlying nonlinear mapping and accurately predict \(y_{\ell+1}\). Then, the output of linear attention in the Transformer can be calculated as follows:
\begin{align}
   \mA := \mZ + \frac{1}{\ell} \mV \mZ (\mK \mZ)^T (\mQ \mZ),
\end{align}
where $\mK, \mQ, \mV$ are appropriately sized key, query, and value matrices, respectively. When predicting \(y_{\ell+1}\), the relevant output of the attention is $A_{d+1, \ell+1}$. Thus, we consider the prediction of linear Transformer (without MLPs) as 
\begin{align}
    \hat{y}_{linear} := A_{d+1, \ell+1}.
\end{align}
This prediction can be simplified to the following form using the reparameterization technique used by \cite{lu2024incontext, lu2025asymptotic}:
\begin{align}
    \hat{y}_{linear} = \text{vec}(\bGamma)^T \text{vec}(\mH_\mZ),
    \label{eq:linear_transformer}
\end{align}
where $\text{vec}(\cdot)$ denotes the vectorization operation, $\bGamma \in \R^{d \times (d+1)}$ is the parameter matrix formed using the entries of $\mV, \mK, \mQ$ matrices while $\mH_\mZ$ is defined as
\begin{align*}
    \mH_\mZ := \vx_{\ell+1} \begin{bmatrix} \frac{d}{\ell} \sum_{i\leq \ell} y_i \vx_i^T & \frac{1}{\ell} \sum_{i \leq \ell} y_i^2 \end{bmatrix} \in \R^{d \times (d+1)}.
\end{align*}
Note that $\bGamma$ is trained for the linear Transformer case as
\begin{align*}
    \argmin_{\bGamma} \sum_{j=1}^{n} \left( y_{\ell+1}^{j} - \text{vec}(\bGamma)^T \text{vec}(\mH_{\mZ^j}) \right)^2 + \lambda \frac{n}{d} \|\bGamma\|_F^2,  
\end{align*}
where $\lambda$ is the regularization constant, $n/d$ factor next to $\lambda$ is used to keep the regularization meaningful in the asymptotic regime we consider. Here, $\{(\mZ^j, y_{\ell+1}^j)\}_{j=1}^{n}$ denotes the $n$ samples used for the training where $\mZ^j$ is formed by \eqref{eq:embedding_matrix}. 

For the Transformer with a nonlinear MLP, the prediction of the model can be similarly written as
\begin{align}
    \hat{y}_{nonlinear} := \vw^T \sigma(\mF^T \text{vec}(\mH_\mZ)),
    \label{eq:nonlinear_transformer}
\end{align}
where $\sigma : \R \to \R$ is the activation function of the nonlinear MLP. Motivated by the random feature model \cite{rahimi2007random, hu2022universality}, we consider $\mF \in \R^{d(d+1) \times m}$ that is randomly initialized and fixed, while similar to the training above, the parameter vector $\vw \in \R^m$ is trained as
\begin{align*}
    \argmin_{\vw} \sum_{j=1}^{n} \left( y_{\ell+1}^{j} -  \vw^T \sigma(\mF^T \text{vec}(\mH_{\mZ^j})) \right)^2 + \lambda \frac{n}{d} \|\vw\|_2^2.
\end{align*}

Then, we measure the ICL error with
\begin{align}
    \E_{(\bxi, \mZ, y_{\ell+1})}\left[ \left( y_{\ell+1} - \hat{y}\right)^2\right],
    \label{eq:ICL_error}
\end{align}
where $\hat{y}$ refers to \eqref{eq:linear_transformer} for the linear Transformer or it denotes \eqref{eq:nonlinear_transformer} for the Transformer with a nonlinear MLP. 

\begin{figure*}[t]
    \centering
    \begin{subfigure}[b]{0.325\textwidth}
         \centering
        \includegraphics[width=0.99\linewidth]{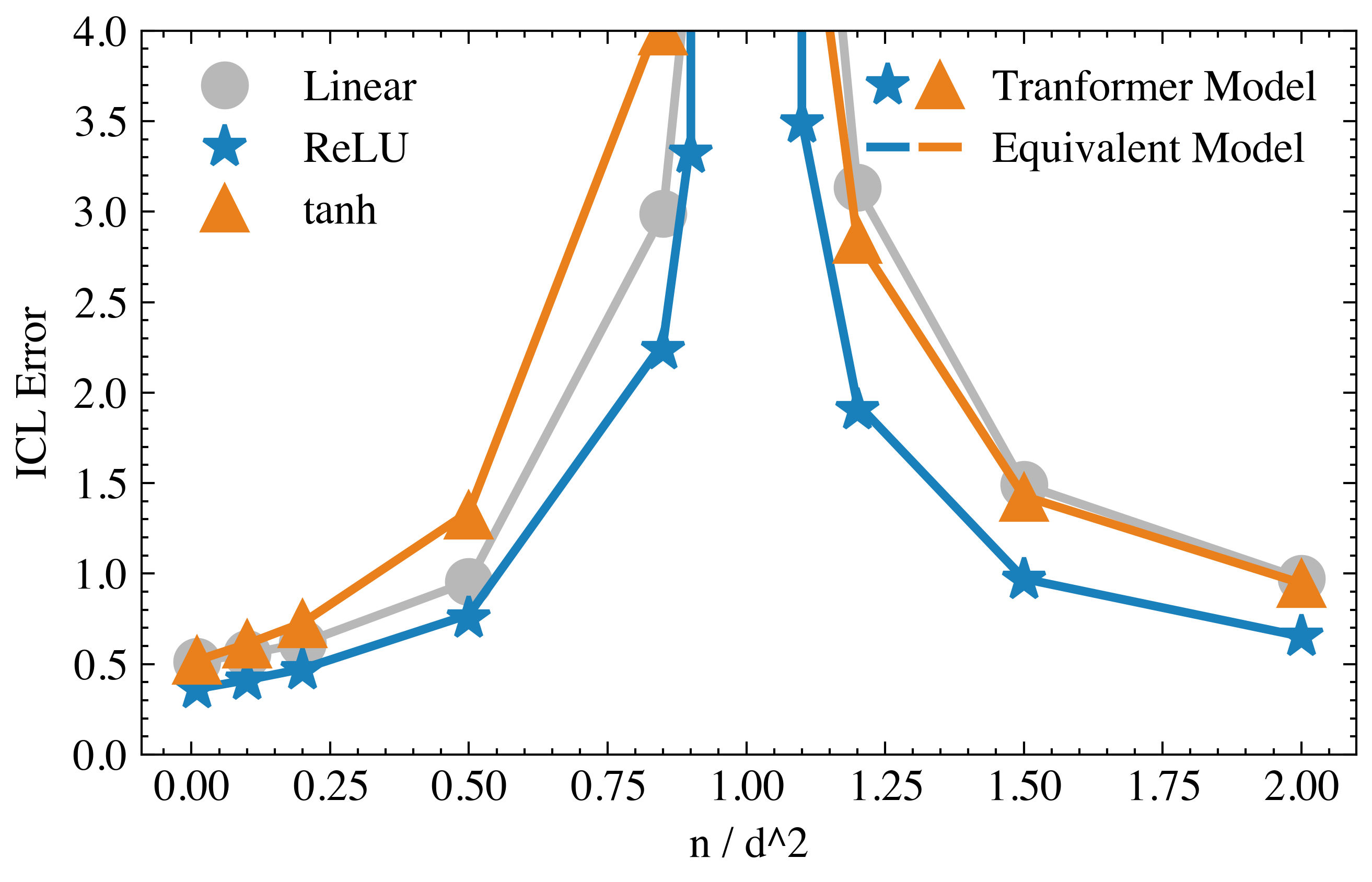}
         \caption{\centering $\sigma_* = ReLU$ }
         \label{fig:impact_of_samples_relu}
     \end{subfigure}
    \hspace{0.175\textwidth}
    \begin{subfigure}[b]{0.325\textwidth}
         \centering
         \includegraphics[width=0.99\linewidth]{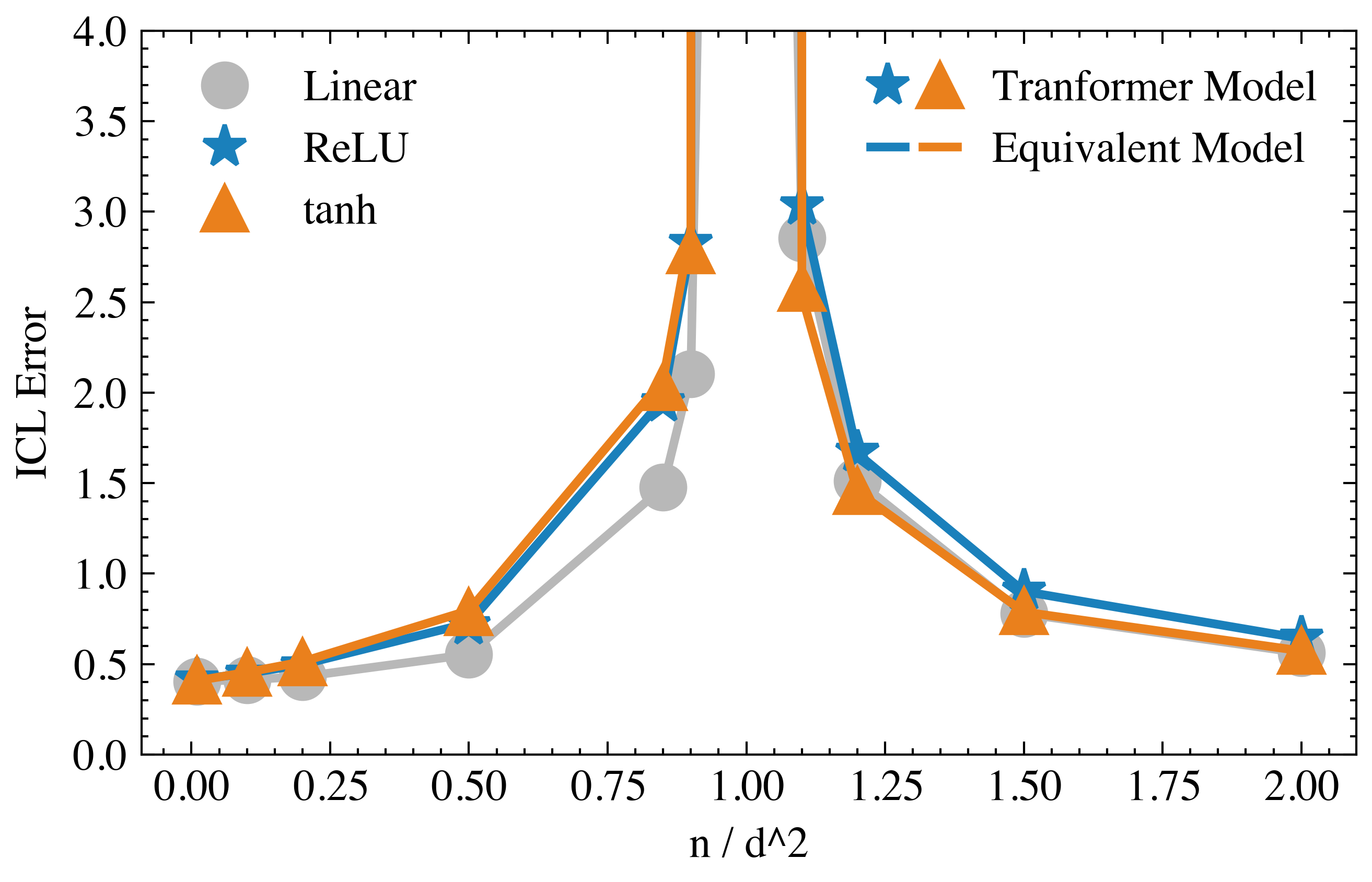}
         \caption{\centering $\sigma_* = tanh$}
         \label{fig:impact_of_samples_tanh}
     \end{subfigure}
     \caption{In-context learning error vs. number of training samples: linear attention model, Transformer, and the equivalent model. The Transformer model is used with two different activation functions (indicated with shapes and colors): ReLU and tanh. The average of 20 Monte Carlo runs is plotted. $n$ is the number of samples, $d=80$, $l=d$, $k = 0.5d$, $m = d^2$, $\rho = 0.01$, $\lambda = 10^{-8}$. }
    \label{fig:impact_of_samples}
\end{figure*}

\section{Main Results}

In this work, we characterize the ICL error \eqref{eq:ICL_error} for the Transformer with a nonlinear MLP in an asymptotic regime. Namely, we assume that the input dimension \( d \), the number of training samples \( n \), the number of task vectors used in training \( k \), the context length \( \ell \), and the hidden dimension \( m \) jointly diverge while $\ell/d, k/d, n/d^2, m/n$ stay constants. Here, the first three scalings have been identified by \cite{lu2024incontext, lu2025asymptotic} for ICL with the linear Transformer, while the last scaling ensures that the model complexity and the training samples are comparable, as assumed typically in the literature \cite{hu2022universality}. 

We start our analysis with an equivalent statistical representation of the random feature mapping $\mF^T \text{vec}(\mH_{\mZ})$ in the following lemma. We later utilize it to analyze the Transformer with a nonlinear MLP \eqref{eq:nonlinear_transformer}.

\begin{lemma}[Asymptotic Distribution of $\mF^T \text{vec}(\mH_\mZ)$]$ $\\
Suppose the entries of $\mF$ are independent and identically distributed as $\mathcal{N}(0, 1/\text{tr}(\text{Cov}( \text{vec}(\mH_{\mZ}) )))$ so that elements of $\mF^T \text{vec}(\mH_{\mZ})$ have unit variance, where $\text{tr}(.)$ and $\text{Cov}(.)$ denote the trace and covariance operators. Let $\vf_i$ denote the $i$-th column of $\mF$. As $l,d \to \infty$ with $l/d \in \R^+$, we have 
\begin{align}
    \vf_i^T \text{vec}(\mH_\mZ) \to \mathcal{N}(0,1) \textit{  almost surely,}
\end{align}
for all $i \in \{1, \dots, m\}$.
\begin{proof}
    Let $t := \text{tr}(\text{Cov}( \text{vec}(\mH_{\mZ}) ))$. For a given $\mH_\mZ$, the conditional distribution of $\vf_i^T \text{vec}(\mH_\mZ)$ is 
    \begin{align}
        \vf_i^T \text{vec}(\mH_\mZ) \mid \mH_\mZ \sim \mathcal{N}(0, \|\text{vec}(\mH_\mZ)\|_2^2/t).
    \end{align}
    Then, by the concentration of $\|\text{vec}(\mH_\mZ)\|_2^2/t$ around $1$, we reach the statement of the lemma.
\end{proof}
\label{lemma:equivalent_representation}
\end{lemma}
This lemma enables us to simplify the term \(\mF^T \text{vec}(\mH_{\mZ})\), facilitating its use in our analysis of in-context learning with Transformers. Consequently, we shift our focus to the joint distribution of \((\mF^T\text{vec}(\mH_{\mZ}),\, \boldsymbol{\xi}^T \vx_{\ell+1})\) rather than \((\mH_{\mZ},\, y_{\ell+1})\), since \(y_{\ell+1} = \sigma_*(\boldsymbol{\xi}^T \vx_{\ell+1}) + \epsilon_{\ell+1}\) by \eqref{eq:data_model}. The following establishes that the pair \((\mF^T\text{vec}(\mH_{\mZ}),\, \boldsymbol{\xi}^T \vx_{\ell+1})\) converges in distribution to a jointly Gaussian random vector in the high-dimensional limit.

\begin{corollary}[Joint Distribution of $(\mF^T\text{vec}(\mH_{\mZ}), \bxi^T \vx_{\ell+1})$]
Suppose that the task vector $\bxi$ is given. Under the assumptions of Lemma \ref{lemma:equivalent_representation}, as $l,d \to \infty$ with $l/d \in \R^+$, $(\mF^T\text{vec}(\mH_{\mZ}), \bxi^T \vx_{\ell+1})$ becomes jointly Gaussian with zero mean and some covariance. Here, the joint distribution appears since the construction of $\text{vec}(\mH_{\mZ})$ includes $\vx_{\ell+1}$.
\label{corollary:asymptotic_gaussianity}
\end{corollary}

Lemmas~1 and Corollary~2 collectively characterize the distributions of both the input \(\text{vec}(\mH_{\mZ})\) of the nonlinear MLP and the label \(y_{\ell+1}\). With these components established, we can now leverage the preceding results in conjunction with existing asymptotic analyses of two-layer neural networks~\cite{demir2024random,demir2025asymptotic} to study the in-context learning error defined in~\eqref{eq:ICL_error} for Transformers equipped with nonlinear MLPs. The following theorem identifies a model that is asymptotically equivalent to such a Transformer architecture, where ``asymptotic equivalence" is defined as both models achieving the same ICL error \eqref{eq:ICL_error} in the asymptotic limit that we consider in this work. 

\begin{theorem}[Equivalent Polynomial Model]
Consider the setting in Lemma \ref{lemma:equivalent_representation}, where $d, n, m, \ell, k$ jointly diverge with $\ell/d, k/d, n/d^2, m/n \in \R^+$. Assume further that \(\sigma\) and \(\sigma_*\) are Lipschitz functions satisfying \(\E_{x \sim \mathcal{N}(0,1)}[\sigma(x)^2] < \infty\), which ensures the existence of a Hermite expansion. Then, the Transformer with a nonlinear MLP  in~\eqref{eq:nonlinear_transformer} is asymptotically equivalent—in terms of ICL error—to the following model: %
\begin{align}
    \vw^T \hat{\sigma}_r(\mF^T \text{vec}(\mH_\mZ)).
\end{align}
Here, $\hat{\sigma}_r : \R \to \R$ is a degree-$r$ polynomial function with a residual term, which is defined as
\begin{align}
    \hat{\sigma}_r(x) := \sum_{i=0}^{r} \frac{1}{i!} c_i He_i(x) + c_r^* z, \quad z \sim \mathcal{N}(0,1),
\end{align}
where $He_i:\R \to \R$ denotes the $i$-th (probabilist's) Hermite polynomial, $c_i$ are the corresponding Hermite coefficients and $c_r^*$ is the residual term such that $\E_{x \sim \mathcal{N}(0,1)}[\hat{\sigma}_r(x)^2] = \E_{x \sim \mathcal{N}(0,1)}[\sigma(x)^2]$. The degree \(r\) depends on both \(\sigma\) and the joint distribution of \((\mF^T \text{vec}(\mH_\mZ),\, \bxi^T \vx_{\ell+1})\), but empirically a small finite \(r\) suffices.

\begin{proof}
The orthogonality of Hermite polynomials with respect to the Gaussian measure can be utilized together with the Corollary \ref{corollary:asymptotic_gaussianity} to prove this theorem, which is omitted here. For an example of such a proof, we refer to \cite{demir2025asymptotic}. 
\end{proof}
\label{theorem:equivalent_model}
\end{theorem}

Theorem~\ref{theorem:equivalent_model} establishes an equivalent yet analytically tractable model for studying random Transformers equipped with a nonlinear MLP for in-context learning. Leveraging this equivalent formulation, we proceed to systematically compare the performance of the Transformer with a nonlinear MLP against that of the linear Transformer. Additionally, we empirically validate the equivalence between the nonlinear MLP Transformer and its polynomial surrogate, highlighting the predictive alignment of the two models under various conditions.

\begin{figure*}[t]
    \centering
    \begin{subfigure}[b]{0.325\textwidth}
         \centering
        \includegraphics[width=0.99\linewidth]{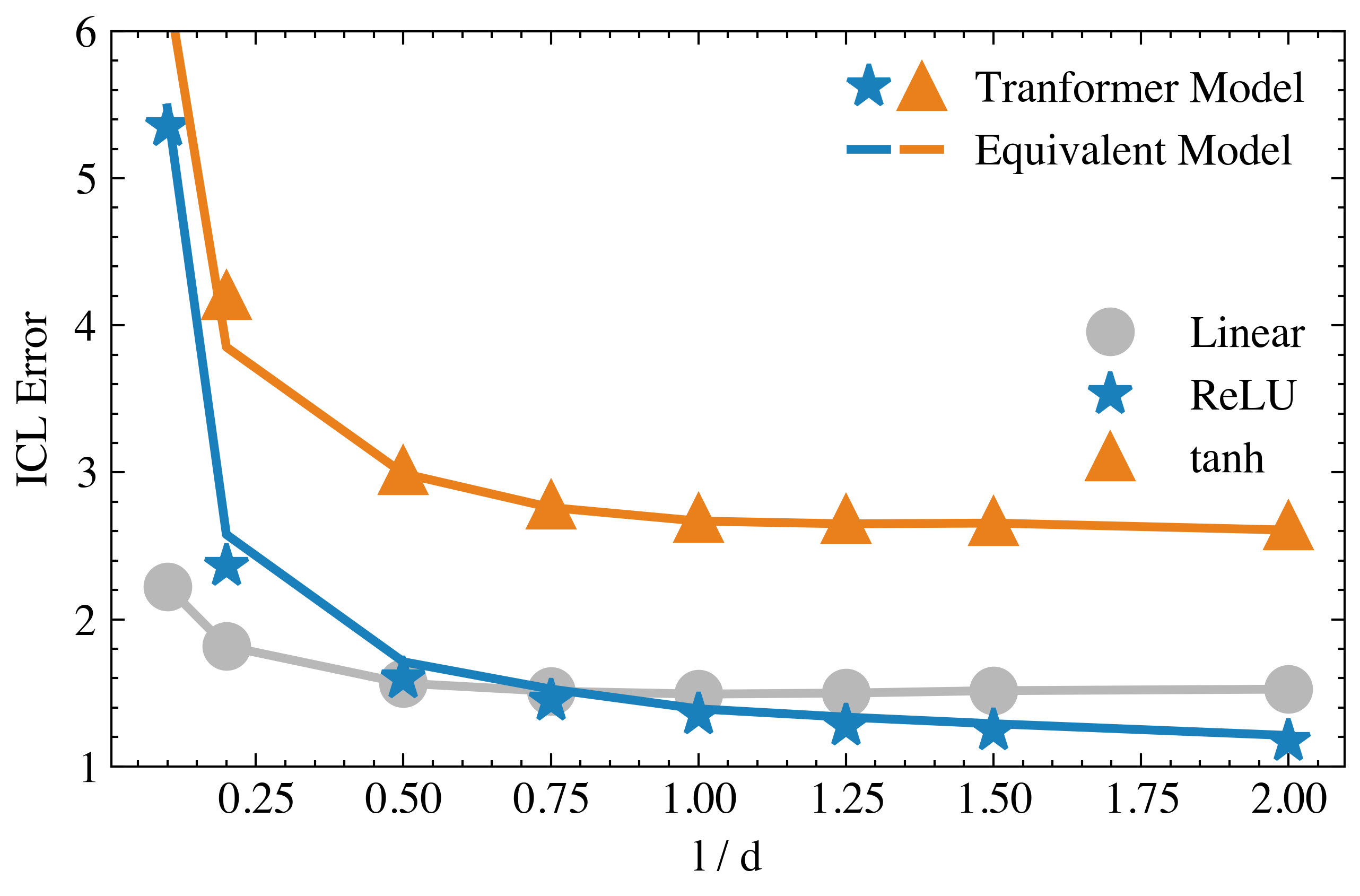}
         \caption{Impact of $\ell$ (with $m = d^2$ and $\lambda=10^{-8}$) }
         \label{fig:impact_l}
     \end{subfigure}
    \begin{subfigure}[b]{0.325\textwidth}
         \centering
         \includegraphics[width=0.99\linewidth]{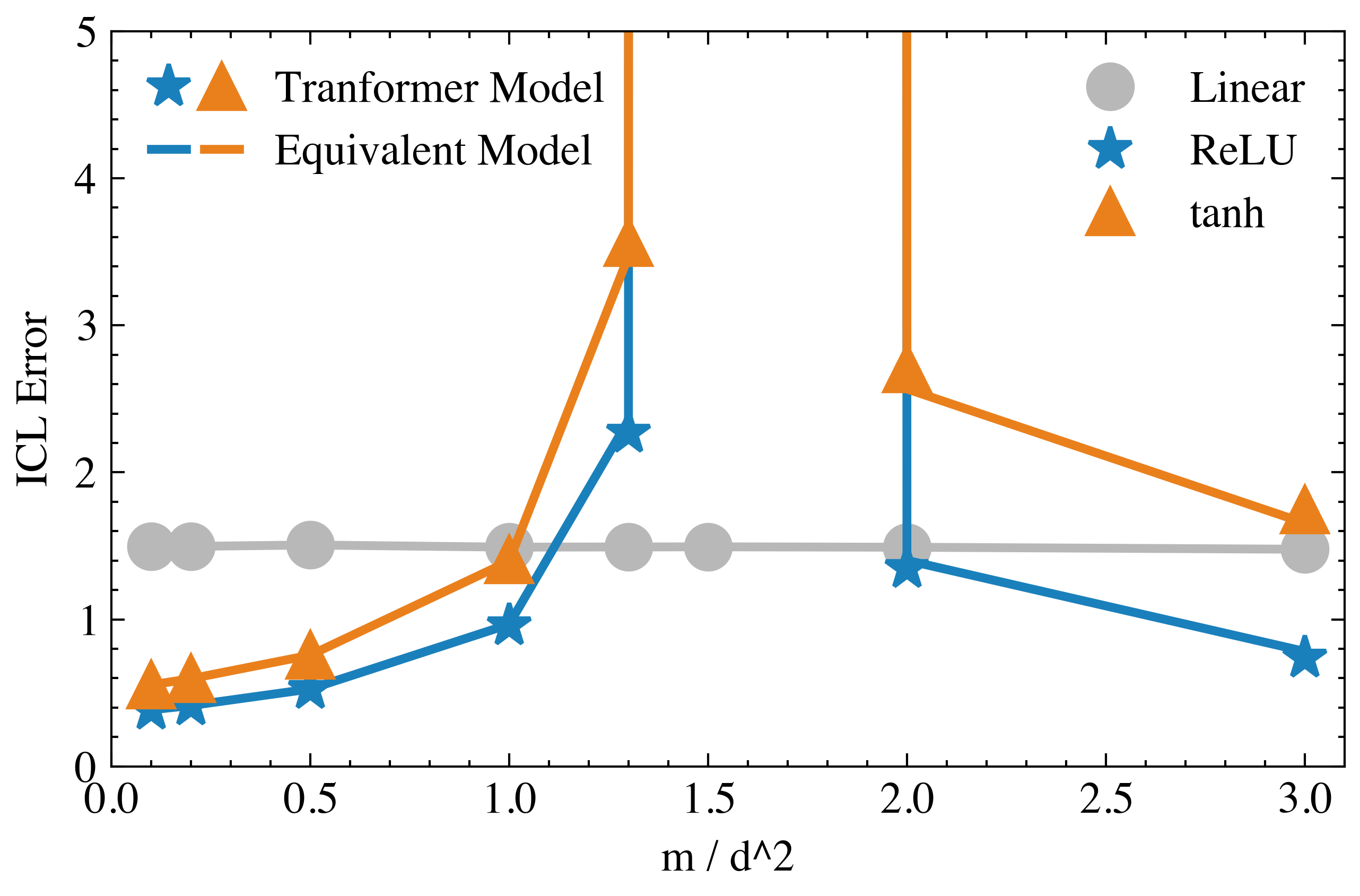}
         \caption{Impact of $m$ (with $\ell=d$ and $\lambda=10^{-8}$)}
         \label{fig:impact_m}
     \end{subfigure}
    \begin{subfigure}[b]{0.325\textwidth}
         \centering
         \includegraphics[width=0.99\linewidth]{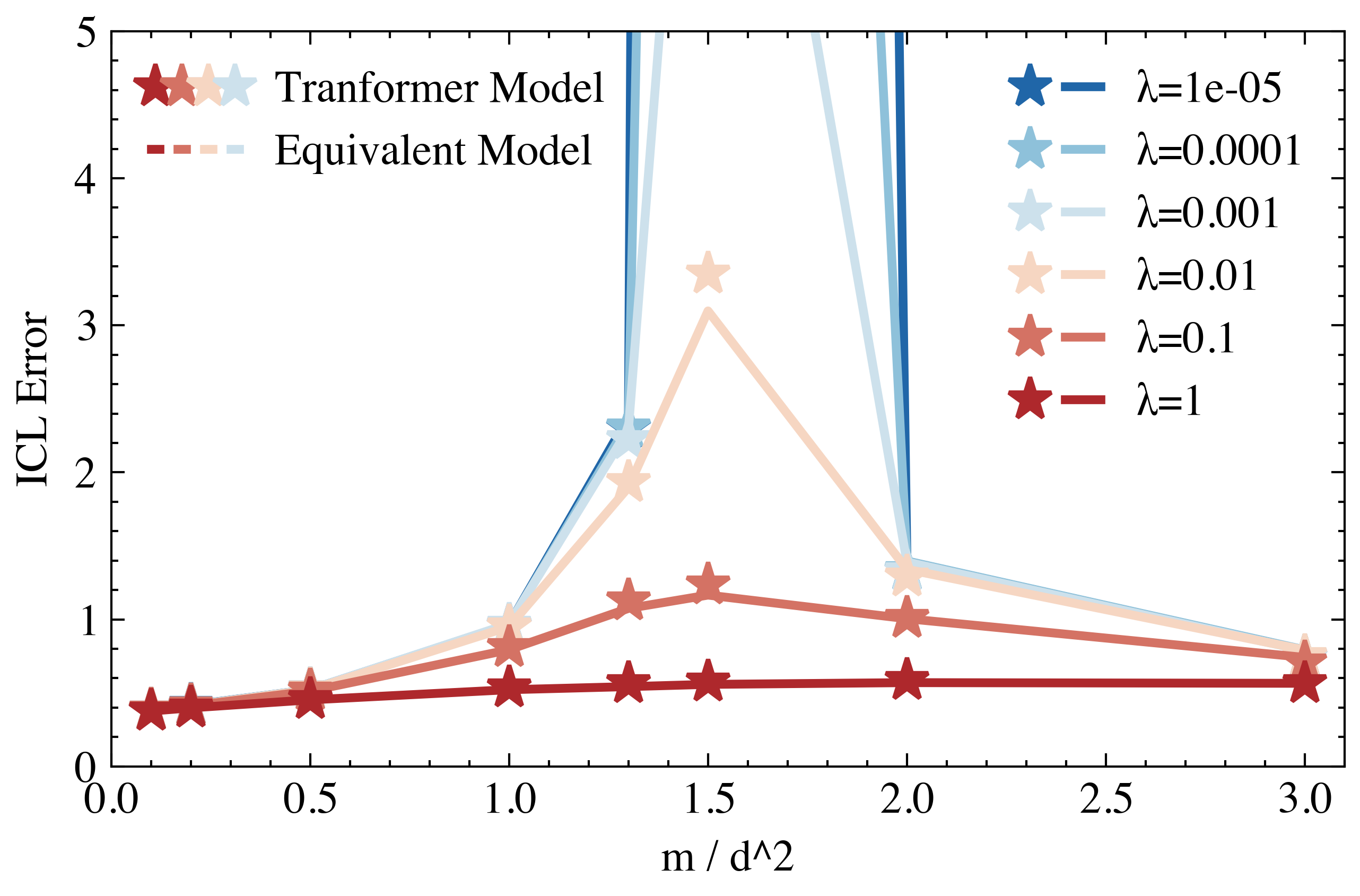}
         \caption{Impact of $\lambda$ (with $\sigma = ReLU$ and $\ell=d$)}
         \label{fig:impact_lambda}
     \end{subfigure}
     \caption{Impact of context length ($\ell$), hidden dimension ($m$), and regularization constant ($\lambda$) on ICL error. In (a), the ICL error gets reduced as we increase the context length $\ell$. In (b), the ICL error exhibits double-descent with respect to the model size, which is mitigated by regularization in (c). For (a)-(b), two different activation functions (indicated with shapes and colors) are employed, while only ReLU activation is used in (c). We illustrate the mean of 20 Monte Carlo runs. Here,  $\sigma_*$ is ReLU, $d=80$, $n = 1.5 d^2$, $k = 0.5d$, $\rho = 0.01$.}
     \label{fig:impact_of_params}
\end{figure*}

Figure~\ref{fig:impact_of_samples} illustrates how the ICL error evolves with the number of training samples $n$. Firstly, a non-monotonic trend is observed due to a relatively small regularization constant. Notably, the ICL errors of the equivalent polynomial model (from Theorem \ref{theorem:equivalent_model}) match those of the Transformer model, confirming our theoretical findings. Furthermore, similar to the case for supervised learning with two-layer neural networks \cite{demir2024random}, ICL performance of the Transformer model is affected by the relationship between the activation function $\sigma$ and the target function $\sigma_*$ based on their Hermite expansions (coefficients), which is illustrated in Figure~\ref{fig:impact_of_samples}a-b. In Figure~\ref {fig:impact_of_samples_relu}, across all values of $n$, the Transformer augmented with a nonlinear MLP head (using ReLU activation) consistently achieves lower ICL error than the linear Transformer baseline. This performance gap underscores the benefit of nonlinear feature processing in this regime. In contrast, the MLP with tanh activation offers limited or no improvement over the linear model, suggesting that the effectiveness of nonlinear heads is highly sensitive to the choice of activation function. When the underlying target function is itself a tanh (Figure~\ref{fig:impact_of_samples_tanh}), the performance gains from using an MLP diminish. The tanh-activated Transformer achieves only marginal improvements over the linear baseline, indicating that alignment between the MLP activation and the true task nonlinearity plays a critical role. These results suggest that MLP layers are most beneficial when their nonlinearities match or complement the structure of the target function.

Figure~\ref{fig:impact_of_params} illustrates the effect of various parameters on the ICL error while confirming that the polynomial model's ICL performance matches that of the random Transformer model. First, in Figure~\ref{fig:impact_l}, we study the effect of the context length $\ell$ on the ICL error, indicating that increasing the context length uniformly reduces the ICL error for all models. Crucially, once the context length exceeds a threshold ($\ell \approx d$ in our setup), the Transformer with an MLP head begins to significantly outperform the linear Transformer. This observation highlights the importance of prompt richness: a sufficiently long context is necessary for the nonlinear MLP components to extract and leverage higher-order statistical structure effectively. Next, in Figure~\ref{fig:impact_m}, the dependence of performance on the hidden dimension $m$ reveals a characteristic non-monotonic behavior commonly known as ``double-descent phenomenon'': the ICL error decreases in the under-parameterized regime, peaks near the interpolation threshold ($m/n \approx 1$), and then declines again as the model enters the over-parameterized regime. This pattern emphasizes the importance of choosing the MLP width carefully. Optimal results are achieved in the suitably over-parameterized regime, where the model can fully exploit nonlinear adaptation capabilities. Finally, regularization (in Figure~\ref{fig:impact_lambda}) mitigates this non-monotonic behavior by smoothing the sharp peak near the interpolation threshold. As the regularization strength $\lambda$ increases, the curve becomes more stable and the error peak is attenuated. These results confirm that proper regularization is essential for mitigating the double-descent phenomenon \cite{nakkiranoptimal2021}.

\section{Conclusion}

In this paper, we presented a high-dimensional analysis of in-context learning (ICL) for Transformers equipped with a nonlinear MLP head. By leveraging tools from Gaussian universality and Hermite polynomial expansions, we proved that a randomly initialized Transformer with an MLP head is asymptotically equivalent to a finite-degree polynomial model. Our theoretical findings precisely delineate the regimes in which nonlinear feature processing provides significant improvements over purely linear attention mechanisms. Extensive simulations support our theory, demonstrating that: (i) the MLP head significantly reduces ICL error when the target function is nonlinear; (ii) the benefits of the MLP emerge only when the context length exceeds certain dimension-dependent thresholds; and (iii) the model’s ICL error exhibits a double-descent behavior with respect to complexity, which can be effectively mitigated through appropriate regularization.

These results not only deepen our theoretical understanding of how nonlinearity and over-parameterization affect ICL but also offer practical guidance for designing Transformer architectures with MLP layers. Future directions include extending the analysis to multi-head attention, exploring deep stacking of MLP blocks, and developing adaptive regularization schemes tailored to the interpolation regime. 

\bibliographystyle{IEEEbib}
\bibliography{refs}

\begin{thebibliography}{10}

\bibitem{vaswani2017attention}
Ashish Vaswani, Noam Shazeer, Niki Parmar, Jakob Uszkoreit, Llion Jones, Aidan~N Gomez, {\L}ukasz Kaiser, and Illia Polosukhin,
\newblock ``Attention is all you need,''
\newblock {\em Advances in neural information processing systems}, 2017.

\bibitem{brown2020language}
Tom Brown, Benjamin Mann, Nick Ryder, Melanie Subbiah, Jared~D Kaplan, Prafulla Dhariwal, Arvind Neelakantan, Pranav Shyam, Girish Sastry, Amanda Askell, et~al.,
\newblock ``Language models are few-shot learners,''
\newblock {\em Advances in neural information processing systems}, 2020.

\bibitem{akyurek2023what}
Ekin Aky{\"u}rek, Dale Schuurmans, Jacob Andreas, Tengyu Ma, and Denny Zhou,
\newblock ``What learning algorithm is in-context learning? investigations with linear models,''
\newblock in {\em International Conference on Learning Representations}, 2023.

\bibitem{pmlr-v202-von-oswald23a}
Johannes Von~Oswald, Eyvind Niklasson, Ettore Randazzo, Joao Sacramento, Alexander Mordvintsev, Andrey Zhmoginov, and Max Vladymyrov,
\newblock ``Transformers learn in-context by gradient descent,''
\newblock in {\em International Conference on Machine Learning}, 2023.

\bibitem{wu2024how}
Jingfeng Wu, Difan Zou, Zixiang Chen, Vladimir Braverman, Quanquan Gu, and Peter Bartlett,
\newblock ``How many pretraining tasks are needed for in-context learning of linear regression?,''
\newblock in {\em International Conference on Learning Representations}, 2024.

\bibitem{zhang2024trained}
Ruiqi Zhang, Spencer Frei, and Peter~L. Bartlett,
\newblock ``Trained transformers learn linear models in-context,''
\newblock {\em Journal of Machine Learning Research}, vol. 25, no. 49, pp. 1--55, 2024.

\bibitem{li2024nonlinear}
Hongkang Li, Meng Wang, Songtao Lu, Xiaodong Cui, and Pin-Yu Chen,
\newblock ``How do nonlinear transformers learn and generalize in in-context learning?,''
\newblock in {\em International Conference on Machine Learning}, 2024.

\bibitem{kim2024transformers}
Juno Kim and Taiji Suzuki,
\newblock ``Transformers learn nonlinear features in context: Nonconvex mean-field dynamics on the attention landscape,''
\newblock in {\em International Conference on Machine Learning}, 2024.

\bibitem{oko2024pretrained}
Kazusato Oko, Yujin Song, Taiji Suzuki, and Denny Wu,
\newblock ``Pretrained transformer efficiently learns low-dimensional target functions in-context,''
\newblock {\em Advances in Neural Information Processing Systems}, 2024.

\bibitem{hu2022universality}
Hong Hu and Yue~M Lu,
\newblock ``Universality laws for high-dimensional learning with random features,''
\newblock {\em IEEE Transactions on Information Theory}, vol. 69, no. 3, pp. 1932--1964, Mar. 2023.

\bibitem{demir2024random}
Samet Demir and Zafer Dogan,
\newblock ``Random features outperform linear models: Effect of strong input-label correlation in spiked covariance data,''
\newblock {\em arXiv preprint arXiv:2409.20250}, 2024.

\bibitem{demir2025asymptotic}
Samet Demir and Zafer Dogan,
\newblock ``Asymptotic analysis of two-layer neural networks after one gradient step under gaussian mixtures data with structure,''
\newblock in {\em International Conference on Learning Representations}, 2025.

\bibitem{nakkiranoptimal2021}
Preetum Nakkiran, Prayaag Venkat, Sham~M Kakade, and Tengyu Ma,
\newblock ``Optimal regularization can mitigate double descent,''
\newblock in {\em International Conference on Learning Representations}, 2021.

\bibitem{demir2023optimal}
Samet Demir and Zafer Do{\u{g}}an,
\newblock ``Optimal nonlinearities improve generalization performance of random features,''
\newblock in {\em Asian Conference on Machine Learning}, 2024.

\bibitem{wei2022emergent}
Jason Wei, Yi~Tay, Rishi Bommasani, Colin Raffel, Barret Zoph, Sebastian Borgeaud, Dani Yogatama, Maarten Bosma, Denny Zhou, Donald Metzler, Ed~H. Chi, Tatsunori Hashimoto, Oriol Vinyals, Percy Liang, Jeff Dean, and William Fedus,
\newblock ``Emergent abilities of large language models,''
\newblock {\em Transactions on Machine Learning Research}, 2022.

\bibitem{olsson2022context}
Catherine Olsson, Nelson Elhage, Neel Nanda, Nicholas Joseph, Nova DasSarma, Tom Henighan, Ben Mann, Amanda Askell, Yuntao Bai, Anna Chen, Tom Conerly, Dawn Drain, Deep Ganguli, Zac Hatfield-Dodds, Danny Hernandez, Scott Johnston, Andy Jones, Jackson Kernion, Liane Lovitt, Kamal Ndousse, Dario Amodei, Tom Brown, Jack Clark, Jared Kaplan, Sam McCandlish, and Chris Olah,
\newblock ``In-context learning and induction heads,''
\newblock {\em Transformer Circuits Thread}, 2022.

\bibitem{schaeffer2023are}
Rylan Schaeffer, Brando Miranda, and Sanmi Koyejo,
\newblock ``Are emergent abilities of large language models a mirage?,''
\newblock in {\em Advances in Neural Information Processing Systems}, 2023.

\bibitem{garg2022can}
Shivam Garg, Dimitris Tsipras, Percy~S Liang, and Gregory Valiant,
\newblock ``What can transformers learn in-context? a case study of simple function classes,''
\newblock {\em Advances in Neural Information Processing Systems}, 2022.

\bibitem{raventos2024pretraining}
Allan Ravent{\'o}s, Mansheej Paul, Feng Chen, and Surya Ganguli,
\newblock ``Pretraining task diversity and the emergence of non-bayesian in-context learning for regression,''
\newblock {\em Advances in Neural Information Processing Systems}, 2024.

\bibitem{bai2023transformers}
Yu~Bai, Fan Chen, Huan Wang, Caiming Xiong, and Song Mei,
\newblock ``Transformers as statisticians: Provable in-context learning with in-context algorithm selection,''
\newblock in {\em Advances in Neural Information Processing Systems}, 2023.

\bibitem{li2023transformers}
Yingcong Li, Muhammed~Emrullah Ildiz, Dimitris Papailiopoulos, and Samet Oymak,
\newblock ``Transformers as algorithms: Generalization and stability in in-context learning,''
\newblock in {\em International Conference on Machine Learning}, 2023.

\bibitem{ahn2023transformers}
Kwangjun Ahn, Xiang Cheng, Hadi Daneshmand, and Suvrit Sra,
\newblock ``Transformers learn to implement preconditioned gradient descent for in-context learning,''
\newblock in {\em Advances in Neural Information Processing Systems}, 2023.

\bibitem{mahankali2024one}
Arvind~V. Mahankali, Tatsunori Hashimoto, and Tengyu Ma,
\newblock ``One step of gradient descent is provably the optimal in-context learner with one layer of linear self-attention,''
\newblock in {\em International Conference on Learning Representations}, 2024.

\bibitem{fu2024transformers}
Deqing Fu, Tianqi CHEN, Robin Jia, and Vatsal Sharan,
\newblock ``Transformers learn higher-order optimization methods for in-context learning: A study with linear models,'' 2024.

\bibitem{li2024finegrained}
Yingcong Li, Ankit~Singh Rawat, and Samet Oymak,
\newblock ``Fine-grained analysis of in-context linear estimation: Data, architecture, and beyond,''
\newblock in {\em Advances in Neural Information Processing Systems}, 2024.

\bibitem{park2024competition}
Core~Francisco Park, Ekdeep~Singh Lubana, Itamar Pres, and Hidenori Tanaka,
\newblock ``Competition dynamics shape algorithmic phases of in-context learning,''
\newblock {\em arXiv preprint arXiv:2412.01003}, 2024.

\bibitem{lu2024incontext}
Yue~M Lu, Mary~I Letey, Jacob~A Zavatone-Veth, Anindita Maiti, and Cengiz Pehlevan,
\newblock ``In-context learning by linear attention: Exact asymptotics and experiments,''
\newblock in {\em NeurIPS 2024 Workshop on Mathematics of Modern Machine Learning}, 2024.

\bibitem{lu2025asymptotic}
Yue~M Lu, Mary Letey, Jacob~A Zavatone-Veth, Anindita Maiti, and Cengiz Pehlevan,
\newblock ``Asymptotic theory of in-context learning by linear attention,''
\newblock {\em Proceedings of the National Academy of Sciences}, vol. 122, no. 28, pp. e2502599122, 2025.

\bibitem{rahimi2007random}
Ali Rahimi and Benjamin Recht,
\newblock ``Random features for large-scale kernel machines,''
\newblock in {\em Advances in Neural Information Processing Systems}, 2007.

\end{thebibliography}

\end{document}